\documentclass{Interspeech2024}
\usepackage{pgf-pie}  
\usepackage{tikz}     
\usepackage{verbatim} 
\usepackage{booktabs} 
\usepackage{tabularx} 
\usepackage{multirow} 
\usepackage{caption} 
\usepackage{pifont}
\usepackage{amssymb}
\usepackage{xcolor}
\usepackage{hyperref}
\usepackage{booktabs}

\definecolor{tableheader}{RGB}{32,78,140}
\definecolor{tableline}{RGB}{221,221,221}

\interspeechcameraready

\title{Exploring Multilingual Unseen Speaker Emotion Recognition: Leveraging Co-Attention Cues in Multitask Learning}
\vspace{-5em}
\name[affiliation={1{^\dagger}{^\#}}]{Arnav}{Goel}
\name[affiliation={1^\dagger}]{Medha}{Hira} 
\name[affiliation={1,2}]{Anubha}{Gupta}
\address{
  $^1$SBILab, Indraprastha Institute of Information Technology Delhi (IIIT-D), India \\
  $^2$MIRAE AI Systems Pvt. Ltd.
  }
\email{arnav21519,medha21265,anubha\{@iiitd.ac.in\}\thanks{$^\dagger$Equal contribution; $^\#$Corresponding Author}}

\begin{document}
\maketitle
\vspace{-5mm}
\begin{abstract}
Advent of modern deep learning techniques has given rise to advancements in the field of Speech Emotion Recognition (SER). However, most systems prevalent in the field fail to generalize to speakers not seen during training. This study focuses on handling challenges of multilingual SER, specifically on unseen speakers. We introduce CAMuLeNet, a novel architecture leveraging co-attention based fusion and multitask learning to address this problem. Additionally, we benchmark pretrained encoders of Whisper, HuBERT, Wav2Vec2.0, and WavLM using 10-fold leave-speaker-out cross-validation on five existing multilingual benchmark datasets: IEMOCAP, RAVDESS, CREMA-D, EmoDB and CaFE and, release a novel dataset for SER on the Hindi language (BhavVani). CAMuLeNet shows an average improvement of approximately 8\% over all benchmarks on unseen speakers determined by our cross-validation strategy.


   
\end{abstract}

\section{Introduction}

\begin{figure*}[]
    \centering
    \includegraphics[width=\textwidth, trim = 0 20 0 80]{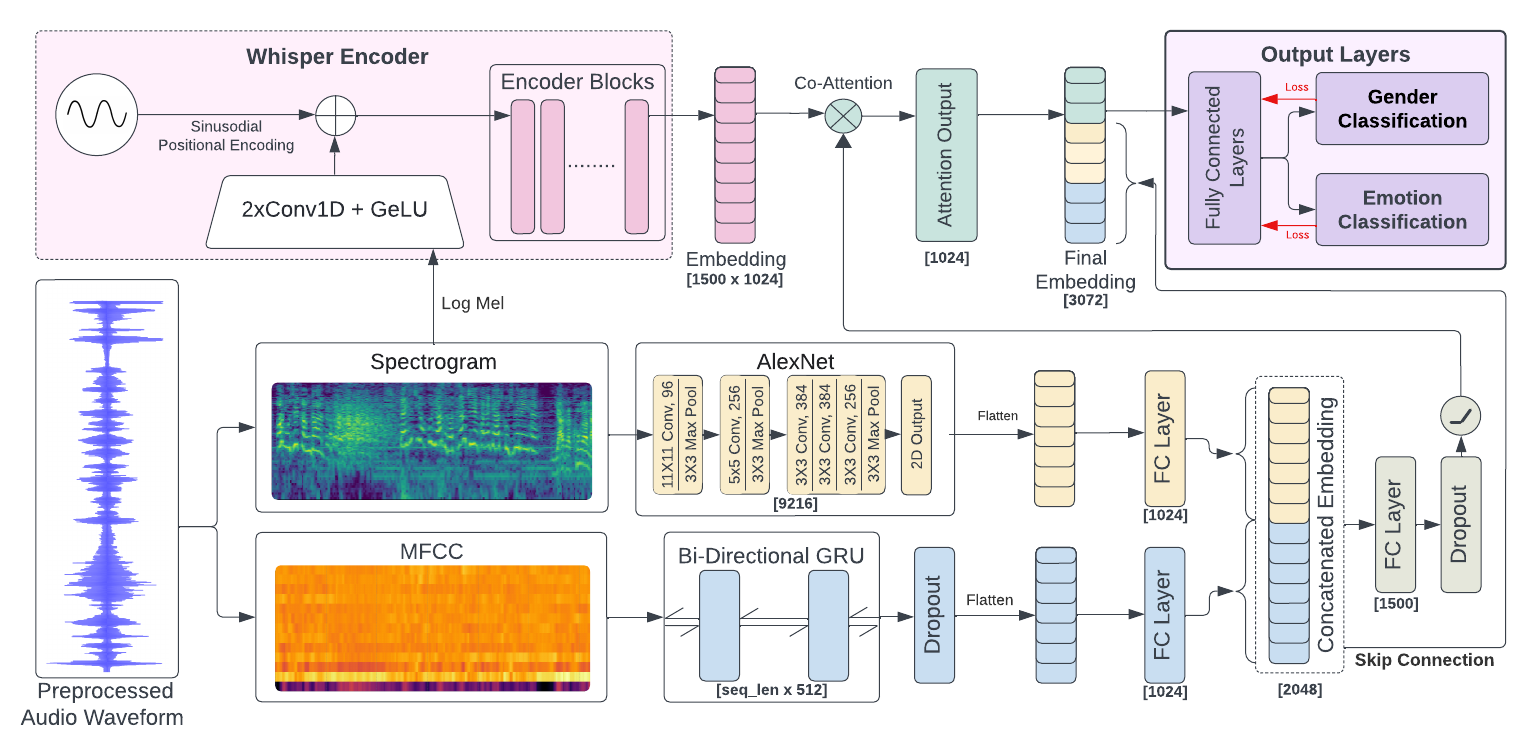}
    \caption{CAMuLeNet: Co-Attention based Multitask Learning Network}
    \vspace{-1em}
    \label{proposed:method}
\end{figure*}

In her seminal work on affective computing, Picard asserts that computers can achieve genuine intelligence and natural interactivity if we empower them with the ability to recognize and understand emotions \cite{picard2000affective}. Besides working with vision-based cues such as facial expressions and hand movements, humans excel in discerning emotions even when only auditory information is available \cite{mayer2008human}. This ability highlights the nuanced and adaptable nature of human emotional understanding, capable of discerning and processing emotions across a wide spectrum of speakers and voice types. Building on this premise, it is imperative to extend the capabilities of Speaker Emotion Recognition (SER) systems beyond speakers on which they have been trained.

Traditional SER systems rely on features including pitch, energy, MFCCs, and spectrograms for emotion recognition \cite{el2011survey, yenigalla18_interspeech}. With the emergence of deep learning methods, systems employing CNNs, Bi-Directional RNNs, and LSTMs are able to learn discernible features \cite{mao2014learning, cho2014learning, lee2015high}. Transformer-based models, a more recent development, marked a significant advancement with the introduction of large pre-trained models (PTMs) trained under a self-supervised learning framework \cite{vaswani2017attention}. Recent advancements in weakly-supervised models, such as Whisper \cite{radford2023robust}, which are trained on extensive corpora, have demonstrated superior performance on a diverse array of downstream tasks \cite{goel2024multilingual, prabhavalkar2023end, hira2024crossvoice}. Attention mechanisms, including cross-attention \cite{he2023multiple}, windowed-attention \cite{chen2023dwformer}, and self-attention \cite{tarantino19_interspeech} along with multitask training have been explored to enhance the performance of SER \cite{9747417, li19n_interspeech, cai21b_interspeech}. Despite these advances, a critical challenge remains: the inability of modern SER systems to effectively adapt to unseen scenarios and speakers, resulting in performance that is inferior to human capabilities \cite{moine2021speaker, antoniou2023designing}. 

This study contributes to the field by benchmarking various PTM embeddings in a transfer learning framework, specifically addressing unseen speaker recognition on five existing benchmark datasets and the 6th newly released dataset with this work. Although co-attention based fusion mechanisms have been used previously on the speaker emotion recognition downstream task for fusing features from multiple modalities \cite{zhao2023knowledge} and multi-level acoustic information \cite{zou2022speech}, their use on unseen speaker emotion recognition tasks along with multitask learning is yet to be thoroughly explored. This study addresses this gap by proposing an architecture that fuses features from the frequency domain and PTM embeddings.

Moreover, the variation in emotional expression across languages poses a distinctive challenge in multilingual SER, which is compounded by the scarcity of comprehensive datasets. To address this gap, we introduce a novel Hindi SER dataset, designed to enhance model training and benchmarking in Indian linguistic contexts. To the best of our knowledge, this is the \textit{first open-source Hindi SER dataset}. Extending our efforts, we apply our methodology to French and German datasets, positioning our work as the first to benchmark Whisper's encoder in multilingual SER settings. The codes and dataset can be found on GitHub\footnote{https://github.com/arnav10goel/CAMuLeNet}.

The key contributions of this work are three-fold:
\begin{enumerate}
    \item We introduce BhavVani, the first-ever Hindi Speech Emotion Recognition dataset with over 9,000 utterances.
    \item Our research uniquely benchmarks pre-trained model embeddings across six datasets in four languages (English, German, French, Hindi) for unseen speaker recognition, marking the first study of its kind to explore these embeddings, including Whisper, on this downstream task.
    \item We introduce \textit{CAMuLeNet}, a Co-Attention based Multitask Learning Network architecture that fuses frequency domain features with PTM features in a multitask framework of emotion and gender recognition, aiming to derive generalized representations for enhanced speaker emotion recognition.
\end{enumerate}
\vspace{-0.5em}

\section{Proposed Methodology} \label{section:method}
\vspace{-0.5em}
The proposed CAMuLeNet architecture described in Figure\ref{proposed:method}, aims to fuse traditional frequency domain features of the spectrogram and MFCCs, with the features extracted from a pre-trained Whisper encoder using the co-attention mechanism described next. We train this architecture through a multitask setup to improve performance on unseen speakers' emotion recognition. 
\vspace{-0.5em}
\subsection{Extracting Features from the Frequency Domain}
\vspace{-0.5em}
To capture frequency and pitch variations in audio clips for generalizing unseen speakers, we utilize frequency-domain features. The spectrogram and mel-frequency cepstrum coefficients (MFCCs) are represented as \(x_{s}\) and \(x_{m}\), respectively, choosing \(40\) MFCCs. Both are used in their two-dimensional form, capturing time-frequency representation without average pooling. An audio clip \(x\) is preprocessed through padding and filtering for consistent sequence length, followed by feature extraction via an AlexNet encoder \cite{krizhevsky2012imagenet}, treating the spectrogram as an image. The latent embeddings from AlexNet is a one-dimensional feature vector \(x'_{s}\) of size \(4096\). Concurrently, MFCCs undergo processing through a Bidirectional Gated Recurrent Unit (Bi-GRU) with two 256-sized hidden layers and a 0.2 dropout, producing a one-dimensional embedding \(x'_{m}\) sized (seq\_len \(\times 512\)).
\vspace{-0.5em}
\subsection{Extracting Features through Transfer Learning}
\vspace{-0.5em}
OpenAI's Whisper\footnote{https://github.com/openai/whisper}, a multilingual encoder-decoder model, exhibits state-of-the-art performance across various speech-to-text benchmarks. Trained on a vast and diverse audio dataset, it significantly improves speech recognition and translation tasks. We hypothesize that Whisper's encoder produces rich latent representations of audio samples. We pass the preprocessed audio clip \(x\) through the Whisper Encoder (which converts it into a mel spectrogram for processing), obtaining a 2D latent representation \(x_{w} \in \mathbb{R}^{L \times W}\). We refrain from using pooling on this 2D representation to maintain the time-frequency information of these embeddings.
\vspace{-0.5em}
\subsection{Co-Attention based Fusion}
\vspace{-0.5em}
The three embeddings hold vital knowledge representation that we aim to fuse for improved performance. First, the derived embeddings from the spectrogram $x'_{s}$ and MFCC features $x'_{m}$ are passed through a fully connected (FC) layer with the same output dimension:
\vspace{-0.5em}
\begin{equation}
\begin{aligned}
    x_{m_{att}} &= f_{att_m}(x'_{m}) &, \quad & x_{s_{att}} = f_{att_s}(x'_{s})
\end{aligned}
\end{equation}
where $x_{s_{att}} \in \mathbb{R}^{1 \times T}$, $x_{m_{att}} \in \mathbb{R}^{1 \times T}$, and $T (=1024)$ is a hyperparameter. The two transformed one-dimensional embeddings are concatenated to create a one-dimensional vector and passed through another FC layer: 
\begin{equation} \label{3}
x_{sm_{att}} = f_{att_{ms}}(x_{m_{att}} \oplus x_{s_{att}})
\end{equation}
where $x_{sm_{att}} \in \mathbb{R}^{1 \times L}$. The activated concatenated features from (\ref{3}) are sent through a FC layer activated by \emph{ReLU} and sent through a \emph{Layer Norm}. This output is multiplied with Whisper features $x_{w}$ to generate attention-weighted Whisper features 
\begin{equation} \label{4}
x_{w_{att}} = f_{att_{ms}}(x_{sm_{att}} \otimes x_{w}),
\end{equation}
where $x_{w_{att}} \in \mathbb{R}^{1 \times W}$. Attention-weighted features are sent through a 3-layer network with 0.15 dropout, an FC layer with ReLU activation, and a terminal Layer Norm. These processed features, along with activated features from (\ref{3}) sent via a skip-connection, are concatenated to form the network's final embedding for downstream classification. This co-attention mechanism attends frequency domain information with frame-level features from the Whisper encoder.

\vspace{-0.5em}
\subsection{Multitask Learning}
\vspace{-0.5em}
Utilizing multitask learning, our network concurrently trains on emotion and gender recognition, leveraging the significant influence of speaker gender on emotion. This approach aims to develop nuanced latent representations that captures intricate emotion-speaker correlations. We use categorical cross entropy loss \(\mathcal{L}_{cce}\) for multiclass emotion recognition and binary cross entropy loss \(\mathcal{L}_{bce}\) for gender recognition, reflecting the prevalent binary gender annotation. The combined multitask objective is defined as
\begin{equation} \label{eq:loss}
    \mathcal{L} = \alpha \cdot \mathcal{L}_{cce} + \beta \cdot \mathcal{L}_{bce} + \gamma,
\end{equation}
with \(\alpha\) and \(\beta\) as tunable weights for each task and $\gamma$ as a tunable parameter providing stability while training. Experimentally, setting \(\alpha\) three times higher than \(\beta\) effectively balances training across tasks, attributed to the class count ratio and the relative simplicity of emotion recognition, thus imposing a higher penalty for better learning from the latter.
\vspace{-0.5em}
\section{Datasets} \label{dataset}
\subsection{Benchmark Datasets}
\vspace{-0.5em}
This study utilizes five widely-used multilingual datasets (English, German, French) to assess pre-trained model embeddings and our methodology for unseen speaker emotion recognition. They are summarised next. 

The CREMA-D dataset \cite{cao2014crema}, featuring 7,442 clips from 91 voice actors (48 males, 43 females), offers recordings of 12 sentences expressed in 6 emotions in English. The IEMOCAP dataset \cite{busso2008iemocap} includes 10,039 clips from 10 actors (5 males, 5 females) in both scripted and spontaneous conversations, annotated with 4 emotion classes. The RAVDESS \cite{livingstone2018ryerson} dataset comprises 1,440 recordings from 24 actors (12 males, 12 females), each articulating two sentences across 8 emotional states. 
Each recording features one speaker expressing a single emotion, with equal distribution of male and female voices across the dataset. Among these, CREMA-D stands out with the largest variety of unique speakers from various ethnicity and is the largest corpus of speech samples among the existing datasets.

We utilize EmoDB \cite{burkhardt2005database} and CaFE \cite{gournay2018canadian} datasets for German and French languages, respectively. EmoDB includes 535 recordings from 10 actors (5 males, 5 females), featuring 10 sentences in \emph{7 emotional states}. CaFE provides a collection of 936 clips from 12 actors (6 males, 6 females), each expressing 6 sentences across \emph{7 emotions}.
\vspace{-0.5em}
\subsection{Our Novel Hindi SER Dataset: \textit{BhavVani}}
\vspace{-0.5em}
In response to the under-representation of Indic languages in Speech Emotion Recognition (SER) and their absence in multilingual benchmarks such as SERAB \cite{scheidwasser2022serab}, we release BhavVani, a novel dataset tailored for Hindi SER. This initiative aims to enrich the SER research landscape by incorporating the linguistic and cultural diversity of Indian languages, leveraging their morphological richness and unique emotional expressiveness.
\vspace{-0.5em}
\subsubsection{\textbf{Data Collection}}
\vspace{-0.5em}
The BhavVani dataset comprises approximately \emph{13 hours of audio} across \textit{8734} utterances, with an average clip length of 5.08 seconds. The audio clips are curated from the popular Indian sitcom "Sarabhai vs Sarabhai", sourced from prior work that used the show's text for sarcasm detection tasks \cite{kumar2022did}. To our knowledge, BhavVani is the \textit{first open-source Hindi dataset tailored for speaker emotion recognition}. The dataset will be released to the community for further work. Each utterance is a single-speaker dialogue, annotated into one of seven categories: Neutral, Surprise, Enjoyment, Disgust and Anger, Fear, Sadness based off Ekman's seven basic human emotions \cite{ekman1999basic}. 
\vspace{-0.5em}
\subsubsection{\textbf{Data Annotation and Validation}} \label{annotation_details}
\vspace{-0.2em}
The BhavVani dataset was annotated by 18 native Hindi speakers, who received preliminary training to ensure annotation consistency and accuracy. Each annotator evaluated approximately 750 audio clips, identifying the speaker’s gender, name, and perceived emotion while excluding clips with excessive background noise or multiple speakers. To uphold the integrity of the dataset, a rigorous validation procedure was followed involving three independent annotators reviewing each annotation. Annotations that were agreed by all three annotators, were marked as confirmed, while those with disagreements were re-evaluated, if necessary, and an alternative emotion was suggested. This meticulous annotation and validation method, evidenced by a \textit{Fleiss Kappa score of 0.637}, highlights substantial annotator agreement, ensuring the reliability and quality of the dataset.  
\vspace{-0.5em}
\subsubsection{\textbf{Dataset Statistics}}
\begin{figure}[h]
    \centering
    \includegraphics[scale = 0.17, trim=60 40 20 40]{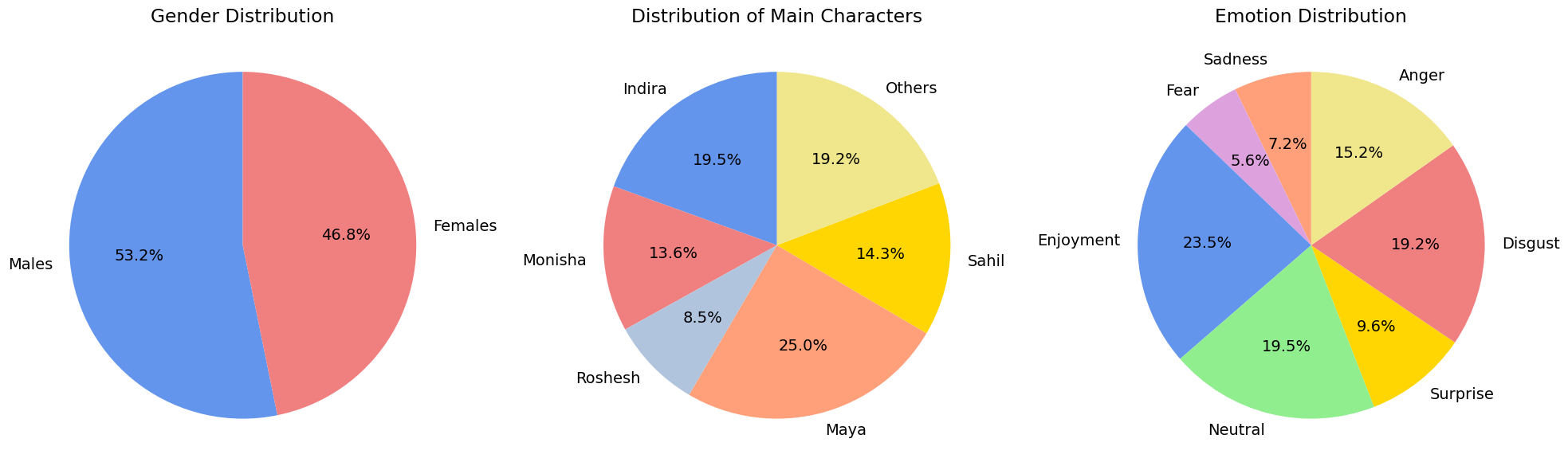}
    \caption{BhavVani Dataset Statistics}
    \label{data_stats}
    \vspace{-1em}
\end{figure}

\vspace{-0.5em}
The pie-charts in Figure \ref{data_stats} provide insights of the thoroughly annotated and curated dataset that had no missing labels. The dataset is balanced with male and female speakers. The gender has been annotated for all speakers, while the speaker name has been annotated primarily for the main characters. None of the other characters had an individual occurrence greater than 41, hence they were positioned under the umbrella of \textit{`Others'}.
\vspace{-0.5em}
\section{Experiments} \label{expts_section}
\vspace{-0.5em}
\subsection{Baseline Pre-Trained Models Employed} \label{baseline_ptm} 
\vspace{-0.5em}
We employed encoders from pre-trained models (PTMs) such as Wav2Vec2.0 \cite{baevski2020wav2vec}, WavLM \cite{chen2022wavlm}, HuBERT \cite{hsu2021hubert}, and Whisper \cite{radford2023robust} as baselines via transfer learning. Initially trained for ASR tasks, Wav2Vec2.0, WavLM, and HuBERT are transformer-based models with 95M parameters each trained through self-supervised learning on English datasets. We fine-tuned these models for multilingual SER using CTC loss. Whisper, distinguished as an encoder-decoder model, is trained with weak supervision on a diverse 680000-hour multilingual dataset, with our focus on its Base (74M) and Medium (769M) variants. We adapt Whisper models for SER by converting their 2D time-frequency outputs into 1D embeddings via average pool. Chosen for their excellence in speech processing, their potential with unseen speakers is yet to be fully explored. 
\vspace{-0.5em}
\begin{table*}[ht]
\vspace{-1em}
\centering
\caption{Comparison of Baseline Methods and CAMuLeNet across various datasets}
\vspace{-1em}
\begin{small}
    \begin{tabularx}{\textwidth}{|l|*{12}{>{\centering\arraybackslash}X|}}
\hline
\multirow{2}{*}{\textbf{PTM}} & \multicolumn{2}{c|}{\textbf{CREMA-D}} & \multicolumn{2}{c|}{\textbf{IEMOCAP}} & \multicolumn{2}{c|}{\textbf{RAVDESS}} & \multicolumn{2}{c|}{\textbf{EmoDB}} & \multicolumn{2}{c|}{\textbf{CaFE}} & \multicolumn{2}{c|}{\textbf{BhavVani}} \\ 
\cline{2-13} 
& WA ($\uparrow$) & WF1 ($\uparrow$) & WA ($\uparrow$) & WF1 ($\uparrow$) & WA ($\uparrow$) & WF1 ($\uparrow$) & WA ($\uparrow$) & WF1 ($\uparrow$) & WA ($\uparrow$) & WF1 ($\uparrow$) & WA ($\uparrow$) & WF1 ($\uparrow$)\\ \hline
Wav2Vec2.0  & 0.451 & 0.409 & 0.427 & 0.408 & 0.396 & 0.377 & 0.469 & 0.455 & 0.417 & 0.372 & 0.251 & 0.317 \\
HuBERT & 0.529 & 0.499 & 0.461 & 0.458 & 0.587 & 0.546 & 0.486 & 0.479 & 0.424 & 0.397 & 0.253 & 0.332 \\
WavLM  & 0.587 & 0.547 & 0.512 & 0.475 & 0.654 & 0.612 & 0.806 & 0.781 & 0.444 & 0.402 & 0.303 & 0.305 \\
Whisper-Base  & 0.603 & 0.591 & 0.573 & 0.588 & 0.593 & 0.587 & 0.743 & 0.732 & 0.452 & 0.441 & 0.261 & 0.265 \\
Whisper-Medium & \textbf{0.666} & \textbf{0.649}& \textbf{0.648} &\textbf{0.667} & \textbf{0.713} & \textbf{0.735} & \textbf{0.831} & \textbf{0.799} & \textbf{0.623} &\textbf{0.551} & \textbf{0.412} & \textbf{0.439} \\
\hline
\textbf{Architecture} & \multicolumn{12}{c|} {\textbf{Experiments on Co-Attention and Multitask Learning}} \\
\hline
CAMuLeNet (Ours) & \textbf{0.762} & \textbf{0.768} & \textbf{0.734} & \textbf{0.728} & \textbf{0.823} & \textbf{0.826} & \textbf{0.862} & \textbf{0.847} & \textbf{0.709} & \textbf{0.691} & \textbf{0.453} & \textbf{0.441} \\
\cline{2-13}
{} & \multicolumn{12}{c|} {\textbf{Ablation Study}} \\
\cline{2-13}
Ours (w/o Multitask) & 0.719 & 0.721 & 0.703 & 0.697 & 0.782 & 0.784 & 0.853 & 0.844& 0.672 & 0.683 & 0.431 & 0.429 \\
Ours (w/o Co-Att \& Multitask) & 0.671 & 0.653 & 0.659 & 0.662 & 0.721 & 0.728 & 0.817 & 0.793 & 0.605 & 0.593 & 0.407 & 0.398 \\
\hline
\end{tabularx}
\end{small}
\textbf{Note:} WA stands for weighted accuracy and WF1 stands for weighted F1 score. These measures account for class imbalance. 
\vspace{-1em}
\end{table*}
\label{table:results}
\subsection{Baseline Experiment Setup} \label{baseline_expt}
\vspace{-0.5em}
In the baseline experiment, we provided 1D-embedding obtained in Section \ref{baseline_ptm} as input to a CNN based feature extractor comprising of a 1D convolutional layer, batch normalization, ReLU activation, dropout (0.3), and max pool followed by flattening and two fully connected (FC) layers for classification. We utilized cross-entropy loss and Adam optimizer with a learning rate of $10^{-4}$ as higher rates led to model overshoot, rapid loss increase, and early underfitting within a few epochs. We trained the models on a NVIDIA A5000 GPU. Training was capped at 20 epochs with early stopping based on validation loss to prevent overfitting. For evaluating performance on unseen speakers, we followed 10-fold leave-speaker-out cross validation, wherein each dataset was segmented into 10 folds with each fold containing unique speakers. Thus, CREMA-D had around 9 new speakers at the time of validation that the model had not seen at the  training time.
\vspace{-0.5em}
\subsection{CAMuLeNet Training Setup}
\vspace{-0.5em}
We extracted MFCC and spectrogram features from the pre-processed audio waveform. The spectrogram is derived by applying a Hamming window based short-time Fourier transform, with a window length of 40, a hop length of 10, setting the size of the FFT window to 800. The calculation of MFCCs involved generating 40 MFCC values with a hop length of 160, ensuring that these coefficients were compatible with the Hidden Markov Model Toolkit (HTK). The training of our model architecture was conducted using a NVIDIA A5000 GPU using batches of 64, leveraging the Adam optimizer with a learning rate of $5\times10^{-5}$. Dropout was maintained at 0.15 throughout the network.  In the context of multitask learning, we determined experimentally that the model achieved optimal stability across different datasets when the weighting factors were set to $\alpha = 0.4$, $\beta = 0.1$ and $\gamma = 0.2$. These values, however, may require tuning to accommodate variations in dataset characteristics and training configurations. Remaining conditions are as described in Section-\ref{baseline_expt}.  
\vspace{-2em}
\section{Results and Discussion}
\vspace{-0.5em}

This Section details our comparative analysis of baseline and proposed methods from \ref{baseline_expt}, quantified by Weighted Accuracy (WA) and Weighted F1 score (WF1) across six datasets, presented in Table 1, complemented by an ablation study discussed later. We report mean metric values averaged over ten folds.
\vspace{-0.5em}
\subsection{Analysis of Baseline Transfer Learning Results}
\vspace{-0.5em}
As per Table-1, Whisper-Medium shows a huge performance jump over self-supervised based PTMs and its base architecture. The increase is noteworthy on the IEMOCAP (13\% over WavLM), CaFE (18\% over Whisper-Base) and BhavVani (14\% over WavLM). The increase is higher on French language dataset due to Whisper being trained on a lot of audio chunks from the French language. Objective performance, however, remains sub-optimal on our Hindi SER dataset indicating the need for interventions to create more robust models and resources for Indic languages.
\vspace{-0.5em}
\subsection{Analysis of Results on CAMuLeNet}
\vspace{-0.5em}
The baseline analysis indicated Whisper-Medium as an optimal pre-trained embedding for our co-attention fusion method. As detailed in Section 4.3, we established a multi-task training framework and observed improved performance across all benchmarks. The introduction of the parameter \(\gamma\) into our loss function contributed to stabilized training, evidenced by a consistent decrease in loss values. Remarkably, the employment of CAMuLeNet yielded a substantial enhancement in accuracy, particularly on the English-language CREMA-D benchmark, \emph{with an increase of 10\%}, and \emph{an 11\% improvement on the RAVDESS benchmark}. Results on CREMA-D are worth noting due to the presence of speakers from various ethnicity uttering emotions at different levels of valence. Hence, Figure-3 contrasts the t-SNE visualizations between Whisper-Medium and CAMuLeNet, with the latter exhibiting pronounced inter-class separation and reduced intra-class variance, which underscores the efficacy of our model in delineating clearer classification boundaries.

Furthermore, the results on multilingual benchmarks, such as CaFE, mirrored the performance gains observed in baseline experiments, underscoring the generalizability of our approach. The multi-task training paradigm notably amplified model's performance on gender recognition tasks, achieving over 95\% accuracy on all benchmarks. These findings confirm the effectiveness of our multi-task and co-attention strategy to improve the model's performance on unseen speakers from various linguistic backgrounds.
\vspace{-0.5em}
\subsection{Ablation Study}
\vspace{-0.5em}
Our proposed methods utilise features from the frequency domain and latent representations generated by a Whisper encoder. Table 1 additionally shows our ablation study by removing the multi-task training setup (1) and then additionally removing co-attention to fuse the features (2). We observed that removing multitask training and replacing it with a single-task training setup reduces performance by an average of approximately 4\% over CAMuLeNet. Replacing co-attention based feature fusion with normal concatenation reduced performance across all benchmark datasets below the best-performing baseline methods by an average of around 10\% over CAMuLeNet. This could be attributed largely to performing concatenation without accounting for the underlying correlations, emphasising the importance of using co-attention based fusion methods.
\begin{figure}[!ht]
\vspace{-1em}
    \centering
    \begin{minipage}{0.23\textwidth} 
        \centering
        \includegraphics[width=\linewidth]{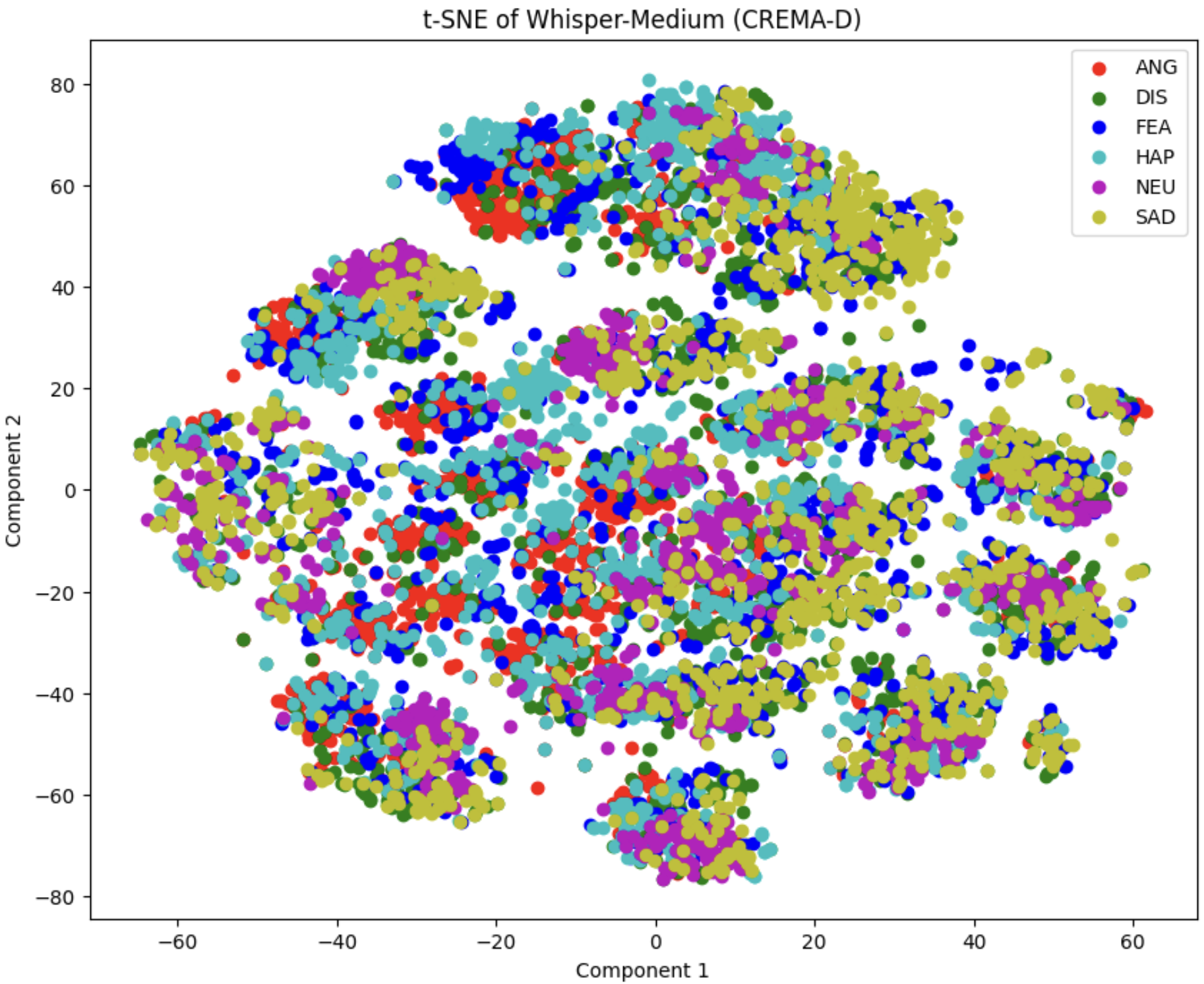} 
        \label{fig:tsne_whisper}
         \textit{\small a) t-SNE for Whisper Med}
    \end{minipage}
    \begin{minipage}{0.23\textwidth} 
        \centering
        \includegraphics[width=\linewidth]{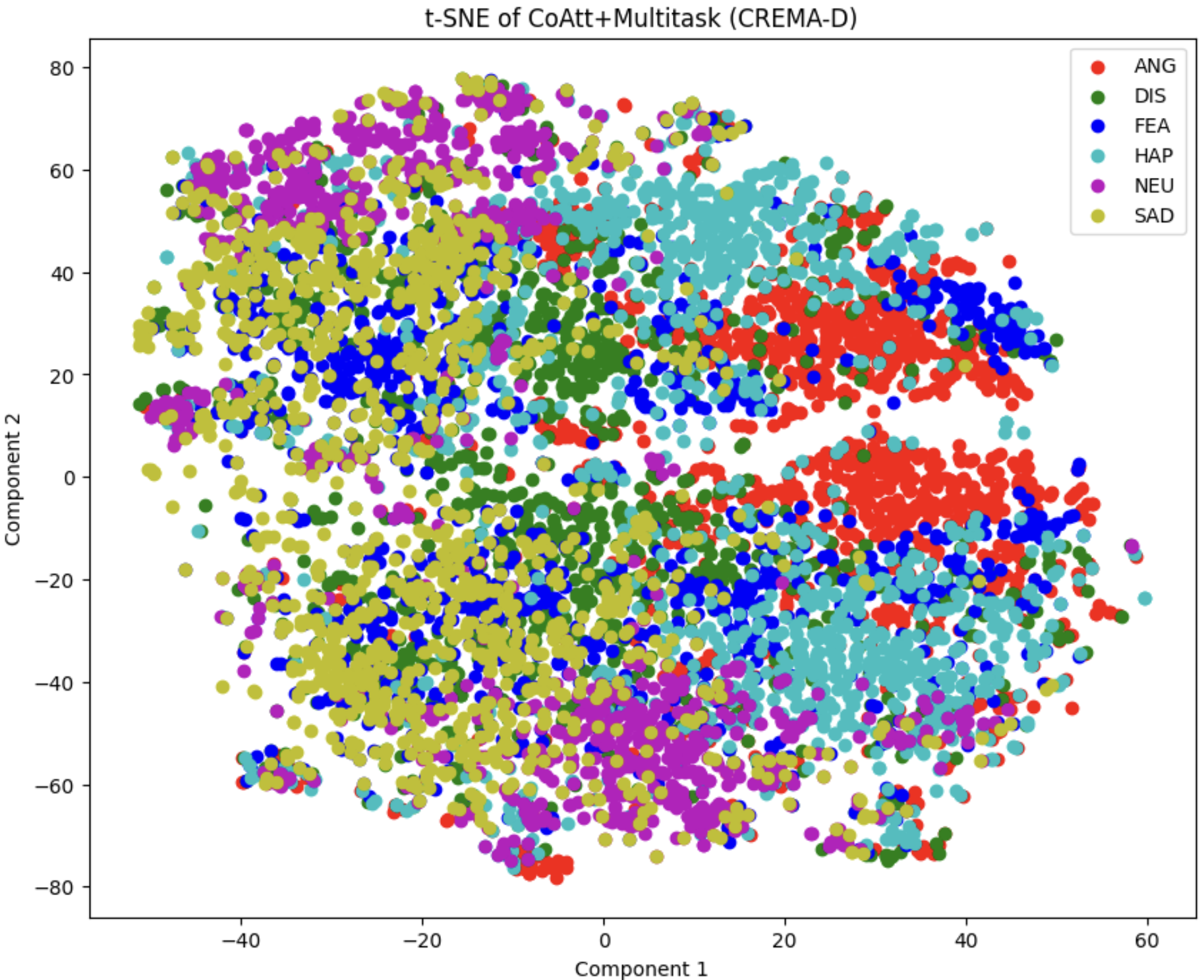} 
        \label{fig:tsne_multi}
        \textit{\small{b) t-SNE for CAMuLeNet}}
    \end{minipage}\hfill 
    \caption{t-SNE visualisation of Feature Distribution. (a) Distribution of final extracted features from Whisper-Medium on the entire CREMA-D dataset. (b) Distribution of final extracted features from CAMuLeNet before the final classification layer on the entire CREMA-D dataset.}
    \vspace{-0.5mm}
\end{figure}
\renewcommand{\arraystretch}{1}
\vspace{-1.5em}
\section{Limitations and Conclusion}
\vspace{-0.5em}
This study exposes a critical limitation in SER for low-resource languages through our novel BhavVani benchmark, where gains were modest and overall performance was subdued, emphasizing the challenges in unseen speaker recognition. Future work will prioritize these languages and investigate alternative fusion mechanisms for robust generalizations. Our contributions include the introduction of BhavVani dataset, comprehensive benchmark of pretrained embeddings across six datasets using a rigorous 10-fold leave-speaker-out cross-validation strategy, and the novel CAMuLeNet architecture, which synergizes frequency domain features with PTM Whisper embeddings through co-attention based feature fusion and multitask training. These efforts spotlight the issue of emotion recognition of unseen speakers for advancing research in this field.

\section{Acknowledgments}
We acknowledge the contributions of: Adamya, Aditya, Akshat, Arnav, Arush, Divyanshi, Kartikay, Kuleen, Manaswi, Manya, Rishitej, Rushil, Sahaj, Tanish, Tanishq and Yash from IIIT Delhi in annotating the BhavVani dataset. This work is supported by the Infosys Foundation via Infosys Centre for AI (CAI), IIIT Delhi. 

\bibliographystyle{IEEEtran}
\bibliography{mybib}

\end{document}